\documentclass[sigconf,natbib=true,anonymous=false,nonacm]{acmart}
\AtBeginDocument{%
  }

\setcopyright{acmlicensed}
\copyrightyear{2026}
\acmYear{2026}
\acmDOI{EvalMG26.SIGIR}
\acmConference[Conference acronym 'SIGIR]{The Second Workshop on Evaluation for Multimodal Generation}{July 24,
  2026}{Melbourne, Australia}
\acmISBN{978-1-4503-XXXX-X/2018/06}

\usepackage{multirow}
\usepackage[export]{adjustbox} 
\usepackage{graphicx}
\usepackage{subcaption}
\usepackage{comment}
\usepackage{algpseudocode}

\begin{document}

\title{Evaluating Visual Language Models on Visual Grounding for Visually Impaired People.}

\author{Fairooz Azim\textsuperscript{1,2}}

\authornote{Work during Orange internship.}
\affiliation{%
  \institution{Orange Research}
   \institution{University of Malta, Msida, Malta}
  \city{Lannion}
  \country{France}
}
\email{fairoozazim97@gmail.com}

\author{Lydia Achour\textsuperscript{1}}
\affiliation{%
  \institution{Orange Research}
  \city{Lannion}
  \country{France}}
\email{lydia.achour@orange.com}

\author{Claudia Borg\textsuperscript{2}}
\affiliation{%
  \institution{University of Malta}
  \city{Msida}
  \country{Malta}}
\email{claudia.borg@um.edu.mt}

\author{Lina Rojas-Barahona\textsuperscript{1}}
\affiliation{%
  \institution{Orange Research}
  \city{Lannion}
  \country{France}}
\email{lina.rojas@orange.com}

\renewcommand{\shortauthors}{Azim et al.}
\newcommand{\todo}[1]{\textcolor{red}{TODO: #1}}
\newcommand{\lr}[1]{\textcolor{blue}{LR: #1}}  
\newcommand{\fa}[1]{\textcolor{magenta}{FA: #1}} 
\newcommand{\la}[1]{\textcolor{green}{LA: #1}} 
\begin{abstract}
  Visual grounding (VG) in visual question answering (VQA) is a fundamental capability of Vision-Language Models (VLMs), enabling the detection of the most relevant region in an image to answer the question. This reflects the model's capability of multimodal comprehension, which is crucial when interacting with visually impaired people (VIP). Indeed, visual grounding, when applied to assistive scenarios, is not merely a technical feature, but rather a trust and safety requirement. In this work, we study visual grounding in VLMs for visually impaired people. To this end, we assess open-source VLMs on visual grounding using the \textsc{VizWiz-VQA-Grounding} dataset and challenge, which contains real-life images taken by VIP. We propose \textsc{VG-SHAP}, which extends an existing explainability framework to provide visual grounding explanations, allowing the evaluation of black-box VLMs. Our results on the \textsc{VizWiz-VQA-Grounding} challenge show that despite generating correct answers, VLMs struggle to ground in the proper region in the image. However, these results are constrained by the quality of the segmentation model of our explainability framework and further research is necessary.
\end{abstract}




\maketitle

\section{Introduction}

Grounding, the collaborative process by which speakers establish and maintain mutual understanding, is crucial in human communication~\cite{CLARK1989259}. \textit{Visual grounding}~\cite{nagaraja2016modeling, yu2016modeling, mao2016generation} in VQA describes a model’s ability to localize the relevant region in the image to answer the question. Despite its importance, recent VLMs tend to answer with self-confidence\cite{shaikh-etal-2024-grounding} rather than grounding their responses. When it comes to assisting visually impaired people (VIP), visual grounding is essential to clearly communicate uncertainties and to correct misunderstandings. Particularly, in critical assistive scenarios, such as verifying the name of a medication, the dosage on a label, or the expiration date on a food package, where the VIP must be able to trust that the VLM’s answer is derived from the correct part of the image and whenever the model is uncertain, it should explicitly convey it. 

VLMs have recently achieved remarkable progress across a wide range of multimodal tasks, particularly in VQA, where models generate natural language responses conditioned on both image and textual inputs~\cite{ribeiro2016should, antol2015vqa}. These advances have enabled practical applications such as assistive systems for visually impaired people (VIPs), where users rely on visual models in their day-to-day tasks~\cite{liu2024right}. One of the most important aspects of this use-case is visual grounding, which will ensure safety and reliability for the VIPs. However, most modern large-scale VLMs are evaluated primarily on textual correctness, and their grounding capabilities on images taken by VIPs remain largely underexplored. Furthermore, many state-of-the-art models operate as black boxes and do not expose internal attention maps, making it difficult to directly assess their visual reasoning processes.


In this work, our contribution is two-fold. First, we extend an existing explainability framework~\cite{cafagna2023} to generate visual grounding explanations, and we integrate recent state-of-the-art segmentation models into it. Second, we assess the visual grounding capabilities of open-source VLMs on the \textsc{VizWiz-VQA-Grounding} dataset and compare them to the baselines reported in the challenge of the same name~\cite{chen2022grounding}. Our preliminary results show that although VLMs tend to generate the correct answer, they do not correctly ground in the proper region in the image, suggesting they lack visual grounding. However, because our evaluation relies on external segmentation models, the results are subject to uncertainty and require further investigation before drawing definitive conclusions about the reliability and trustworthiness of VLMs.



\section{Related Work}
Model-agnostic grounding techniques are paramount for evaluating foundation models as they operate without requiring access to internal weights or specialized fine-tuning, thereby offering unparalleled flexibility and scalability across diverse black-box VLMs. These approaches fundamentally rely on perturbation-based attribution, where segmented image regions are systematically masked or altered to measure their impact on the model's output, such as a prediction score or generated text (\cite{wan2024contrastive}). Among these, Shapley-based explainability methods (e.g.,\cite{cafagna2023, garreau2020explaining}) stand out due to their robustness. They integrate segmentation with a SHAP-based algorithm to quantify the contribution of each image segment to a VLM's black-box output. Crucially, the resulting contribution scores on object segments directly serve as a pixel-level measure of visual grounding, inherently outputting a mask that localizes the most critical visual evidence. In this work, we extend the explainability framework proposed in ~\cite{cafagna2023}, namely V\&L-SHAP. Instead of generating explanations in natural language, we formalize explanations in terms of visual grounding. Moreover, we explore more recent state-of-the-art segmentation models.

In the same spirit of our work, a dataset of egocentric Video QA taken from blind individuals and VIPs was recently released, namely EgoBlind\cite{xiao2025egoblind}. In EgoBlind, $16$ Multimodal Large Language Models were evaluated on QA, performing poorly according to the judgment of GPT-4o mini. Unlike EgoBlind, we focus on visual grounding of static images by utilizing the dataset \textsc{VizWiz-VQA-Grounding} \cite{chen2022grounding}, by evaluating with the objective metrics of the challenge of the same name (introduced in Section~\ref{s:vizwiz}) and by leveraging and extending the model-agnostic explainability framework introduced in Section~\ref{s:expl}.

\section{Visual Grounding for VIP}
\label{s:vizwiz}
An answer that is factually correct but derived from an irrelevant correlation is unusable and potentially dangerous, especially for vulnerable populations, such as VIP. The absence of a verifiable grounding mechanism forces VIP to accept the answer on faith, which is an unacceptable risk for a safe application.
In this work, we use the dataset and challenge \textsc{VizWiz-VQA-Grounding} \cite{chen2022grounding}, created to assist VIP, containing images taken by VIPs, which can be blurry or out of focus, in realistic situations\footnote{\url{https://vizwiz.org/tasks-and-datasets/vqa/}}. This is critical for improving trust, interpretability, and privacy preservation in real-world systems. Since ground truth masks for the test set are withheld, evaluation is conducted through submissions to the EvalAI leaderboard\footnote{\url{https://eval.ai/web/challenges/challenge-page/2182}}.

The metrics used to evaluate visual grounding in this challenge are intersection over union (IoU) and Dice similarity coefficient (Dice Score). IoU measures the overlap between the predicted segmentation mask and the ground truth mask. It is defined as the ratio of the intersection area to the union area of the two masks:
\[
\text{IoU} = \frac{|P \cap G|}{|P \cup G|} \hspace{0.2cm}
\text{Dice} = \frac{2|P \cap G|}{|P| + |G|}
\]
where \(P\) and \(G\) denote the predicted and ground truth masks, respectively. IoU values range from 0 to 1, with higher values indicating better alignment.
The Dice Score is another widely used overlap metric that emphasizes agreement between the predicted and ground truth regions. This metric is more sensitive to small overlaps than IoU and is especially useful in evaluating segmentations with imbalanced class distributions.

\section{Visual Grounding Explainability: VG-SHAP} 
\label{s:expl}
\begin{figure}[ht]
    \centering
    \includegraphics[width=0.82\linewidth]{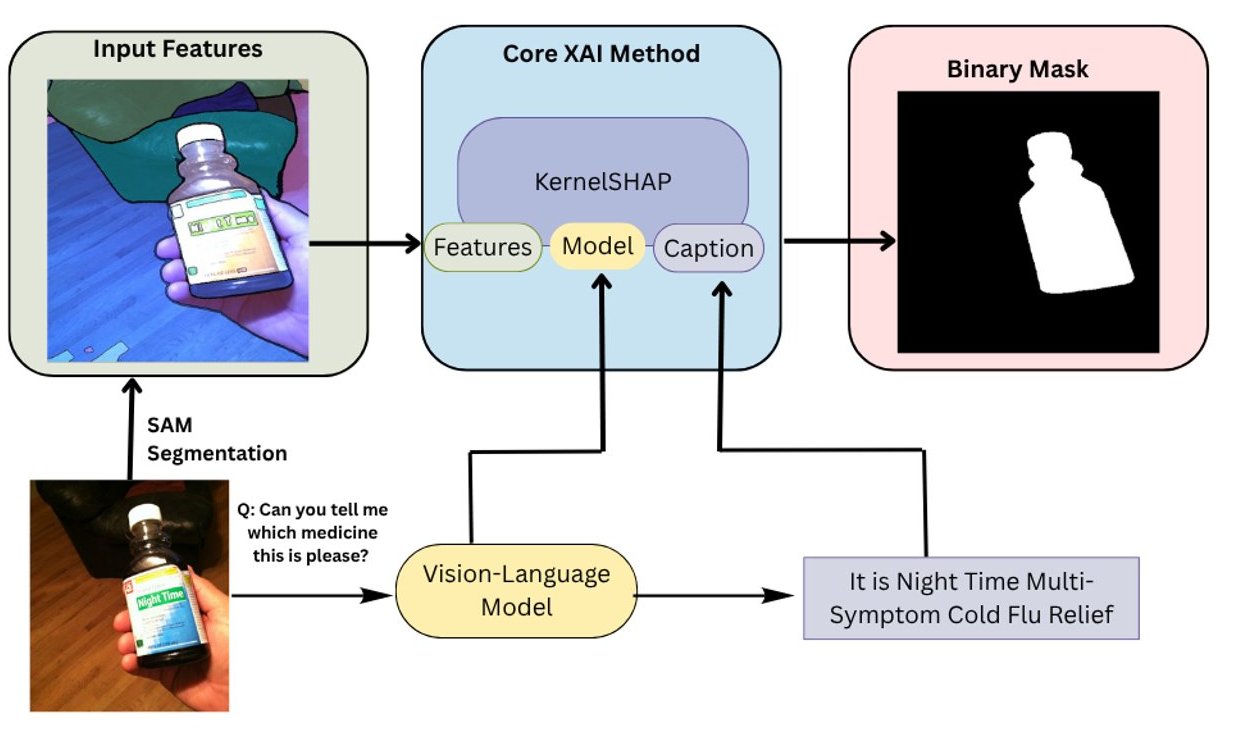}
    \caption{VG-SHAP Explainability Framework.}
    \label{f:explfwk}
\end{figure}

 V\&L-SHAP~\cite{cafagna2023} was originally designed to generate visual explanations in natural language. Instead, we extend it to generate visual grounding from pretrained vision-language models, namely \textbf{VG-SHAP}. Figure~\ref{f:explfwk} depicts VG-SHAP's core explainability method, its inputs and outputs. Feature extraction algorithms (explained in Section~\ref{ss:featureextr}) are used to generate the inputs to the core explainability module (introduced in Section~\ref{ss:shapsai}). Finally, we formalize the output as a binary mask, used to compute visual grounding. 

\paragraph{\textbf{Feature Extraction:}}
\label{ss:featureextr}
this stage involves segmenting an image into meaningful, non-overlapping regions. In the original V\&L-SHAP, CNN-based Deep Feature Factorization (DFF) was used for semantic segmentation, but it produces noisy, inconsistent results on complex real-world VIP images. We therefore replace DFF with: STEGO~\cite{hamilton2022unsupervised}, a self-supervised transformer using energy-based graph optimization; SAM~\cite{kirillov2023sam} and SAM2~\cite{ravi2024sam}, general-purpose prompt-free segmentation models; as well as GroundedSAM2~\cite{ren2024grounded}, which combines SAM2 with Grounding DINO~\cite{liu2024grounding} for semantically grounded, object-level segmentation. After segmenting, the boolean segmentation array is extracted from each segment and converted to a standardized binary mask (dtype uint8), discarding other method-specific metadata. This unified binary masking format ensures that structurally different segmentation outputs, whether class-labeled pixel maps (STEGO) or instance-level masks (SAM variants), are normalized before being passed to the explainer. 


\paragraph{\textbf{Explainer}:}
\label{ss:shapsai}
we leverage the Kernel SHAP~\cite{cafagna2023}, which uses weighted linear regression to estimate the Shapely values. This is computed by evaluating the model across multiple feature subsets \textit{(coalitions)} and averaging contributions over all combinations. The features are image segments, the model is the VLM, and the outcome is the answer to a given question. As in~\cite{cafagna2023}, we used the deterministic sampling approach where the coalitions with higher weights are prioritized, by taking the k-best samples. The core explainer algorithm is as follows. For further details we invite the interested reader to refer to~\cite{cafagna2023}.



\noindent Given a VLM $f$, an image $x$, a question $q$ and a set of $n$ coalitions $\mathbf{C}$
\begin{algorithmic}[1]\raggedright
\State \textbf{input:}   $f$, $q$,  $x$,   $\mathbf{C}=\{c_1,...,c_k,..., c_n\}$. 
\State \textbf{output:}  $\mathbf{c^*}$, the k-best coalitions
\State $a_{ref}$ = f($x$,$q$) the reference answer and $e_{ref}=E(a_{ref})$ its embeddings.
\State $\mathbf{C_{samples}}$, sampled coalitions from $\mathbf{C}$
\For{every coalition $c_i$ in $\mathbf{C_{samples}}$} , 
\State $a_i$ = f($c_i$,$q$) and $e_i=E(a_i)$  its embeddings 
\State $s_i=cos(e_ {ref},e_i)$. 
\EndFor
\State $W=\mbox{kernelSHAP}({s_1,...,s_n})$
\State return $c^*= \mbox{k-best(W)}$
\end{algorithmic}
 
\section{Experimental Setup}
We first conduct a preliminary qualitative analysis of the zero-shot performance of state-of-the-art VLMs on VQA for VIP. Then, we explore distinct segmentation models to solve efficacy issues. Finally, we evaluate the best models on the official \textsc{VizWiz-VQA-Grounding} challenge, which requires an online submission since test-set annotations are not publicly available. 


We consider two parameter configurations for VG-SHAP. Each configuration specifies: (i) the number of points per batch, i.e., pixel-level prompts $(x, y)$ processed simultaneously by the segmentation model; (ii) the minimum mask region size, expressed in pixels, which controls the granularity of the segmented regions; and (iii) the kernel SHAP sampling parameter $k$, denoting the number of sampled region combinations used to estimate feature importance. \textbf{P1} uses 32 points per batch, a minimum mask region size of 500 pixels, and kernel SHAP sampling with $k=500$. \textbf{P2} retains the same batch size, reduces the sampling budget to $k=300$ and increases the minimum mask region size to 4000 pixels, resulting in coarser segmentation and lower coalitions. All experiments run in a high-performance computing (HPC) environment with nodes containing heterogeneous GPU capacity (e.g., H100, H200 and RTX 3090 GPUs). Besides the metrics used for visual grounding, introduced in Section~\ref{s:vizwiz}, we also evaluate: \textsc{accuracy}(\cite{antol2015vqa}); \textsc{BLEU} score, which measures n-gram overlap~\cite{gardent-etal-2017-webnlg}; \textsc{ROUGE-1} and \textsc{ROUGE-L}, which evaluate recall-based lexical similarity and \textsc{Fuzzy Score}, which computes the normalized insertion/deletion similarity between two strings (i.e. the Levenshtein distance). 

\paragraph{\textbf{Preliminary inspection on Zero-shot VQA}:} we conducted a preliminary qualitative evaluation of the zero-shot VQA performance of five state-of-the-art Vision-Language Models (VLMs) on 50 randomly selected samples from the training split of the \textsc{VizWiz-VQA-Grounding} dataset. We consider five multimodal (vision-language) models: \textsc{Gemma3-4B}, \textsc{Qwen2.5-3B}, \textsc{Molmo-7B}, \textsc{LLaVA-7B}, and \textsc{Phi-4-multimodal-instruct (5.6B)}\footnote{Note that this preliminary zero-shot analysis on 50 randomly selected samples is intended only for exploratory model selection. The main grounding evaluation reported in Table~\ref{tab:valtst-iou} is conducted on the VizWiz-VQA-Grounding validation set and official challenge benchmark.}.
Each model received only the image and the corresponding question from the dataset. This setup allows the models to respond freely, without prompt-induced constraints, and offers a clearer view of their default behaviour in a zero-shot setting. The qualitative analysis was based on three criteria: descriptiveness (answer completion), correctness (factual grounding) and hallucination level. 
We first experimented with the following segmentation models in VG-SHAP: \textsc{STEGO}~\cite{hamilton2022unsupervised} and SAM~\cite{kirillov2023sam}. These models provide general-purpose, high-quality masks that align well with the needs of SHAP-based attribution. While \textsc{SAM} excels at zero-shot object segmentation with fine-grained detail, \textsc{STEGO} provides class-consistent, semantically coherent segments.


\paragraph{\textbf{Segmentation Models Evaluation}:}
with the goal of increasing efficiency while reducing the number of segments (i.e., coalitions), we further inspect SAM2~\cite{ravi2024sam} and GroundedSAM2~\cite{ren2024grounded}. We compared SAM, SAM2 and GroundedSAM2 as segmentation models for evaluating \textsc{Gemma3-4B} in a 10-fold cross-validation on 50 randomly selected samples from the validation split of the \textsc{VizWiz-VQA-Grounding} dataset. The metrics used in this step were METEOR, an n-gram metric that also considers synonymy and stemming, and \textsc{BERTScore} for semantic similarity. Finally, we also measure execution time and speed on a server equipped with an RTX 3090 (24GB VRAM). We use the standardized assistive prompt reported in Appendix~\ref{app:prompt}, designed to encourage concise and reliable responses for VIP.
\paragraph{\textbf{\textsc{VizWiz-VQA-Grounding} challenge}:}  The two best prompted models of the preliminary steps are then evaluated on the \textsc{VizWiz-VQA-Grounding} validation set and its official challenge test-set.  

\section{Results}

\paragraph{\textbf{Preliminary inspection on Zero-shot VQA}:} All models demonstrated competitive performance in terms of answer completeness. \textsc{Phi-4-multimodal-instruct} occasionally generated excessively long answers or produced overly brief or under-informative responses; and at times, included irrelevant content.  In terms of correctness, \textsc{LLaVA-7B} showed the weakest performance, even for images that were visually clear. The \textsc{Phi-4} model performed moderately well, though with more variability. \textsc{LLaVA-7B} also exhibited significant hallucination issues, particularly when extracting textual or numerical information such as movie titles from DVDs, time from digital clocks, or object counts. In one instance, it generated a descriptive output that was entirely unrelated to the image content. The \textsc{Phi-4} model similarly showed tendencies toward hallucination, such as misidentifying an inhaler as an aerosol spray. In contrast, \textsc{Qwen2.5-3B}, \textsc{Molmo-7B}, and \textsc{Gemma3-4B} generally produced answers of standard length, consistently generated correct, visually grounded answers and demonstrated minimal hallucination across the evaluated samples. For the sake of frugality, we consider the smallest models in our study, whose performance is presented in Table~\ref{tab:zshot-inf}.

\begin{table}[ht!]
    \centering
    \begin{tabular}{l|c|c|c|c|c}
       Models  & Acc. &  \textsc{Bleu} & \textsc{Rouge-1} & \textsc{Rouge-L} & \textsc{Fuzzy}\\
        \hline
        Gemma3-4B  &0.376 & 0.107 &0.535 & 0.533 &83.74\\
        Qwen2.5-VL-3B & \textbf{0.602} &\textbf{0.165} & \textbf{0.675} & \textbf{0.675} & \textbf{87.54}\\
    \end{tabular}
    \caption{Zero-shot VQA evaluation (Acc. means accuracy).}
    \label{tab:zshot-inf}
\end{table}
\begin{figure}[h!] 
 \begin{tabular}{cc}
  \begin{subfigure}[]{0.15\textwidth}
    \includegraphics[scale=0.15]{images/VizWiz_val_00000315_resized.png}
    \caption{}
    \label{fig:main_image}
    \end{subfigure}&
  \begin{subfigure}[]{0.23\textwidth}
    \includegraphics[width=\textwidth]{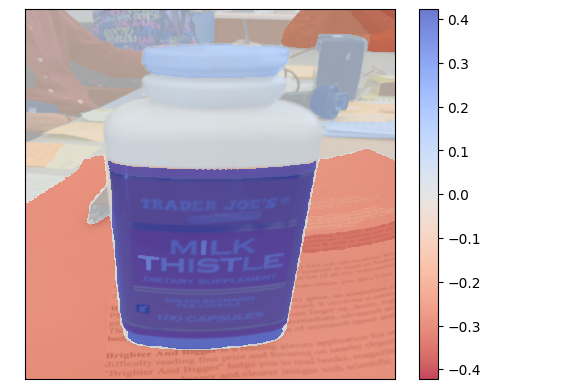}
    \caption{}
    \label{fig:sam_explanation}
  \end{subfigure}
 \end{tabular}
  
\caption{VG explanation (Gemma-3-4B and SAM). (a) original image and (b) binary mask explanation, blue and red mean positive and negative impact on the answer, respectively.}
    \label{fig:sam-ex}
\end{figure}

\paragraph{\textbf{Segmentation Models Evaluation}:} a \textit{preliminary study} on the impact of the segmentation models on a 10-fold cross-validation over 50 randomly selected samples of the validation set of \textsc{VizWiz-VQA-Grounding} is presented in Table~\ref{tab:segmentation}. We observe that Grounding-SAM is remarkably faster and more accurate than \textsc{SAM} and \textsc{SAM2}. It achieves the best segmentation quality ($+0.1$ IoU) compared to \textsc{SAM} and ($+0.08$ IoU) compared to \textsc{SAM2}. Adapting \textsc{Grounding-SAM} will clearly reduce the computational cost at inference time. For a fair comparison, we report the time and speed for a single fold run on a 4-GPU server with an RTX 3090 (24GB VRAM), since we were unable to enforce the same resource allocation conditions on the HPC cluster.

\begin{table}[h!]
    \begin{tabular}{l|l|l|l|l|l|l}
Model & \scriptsize{IoU} & \scriptsize{Dice} & \scriptsize{METEOR} & \scriptsize{BERT} & \scriptsize{$t$} &  \scriptsize{$v$} \\\hline
\textsc{SAM} & $0.35\pm 0.4$ & $0.40\pm 0.40$ & $0.28$ & $0.74$ & $288$ & $1$x\\
\textsc{SAM2} & $0.37\pm 0.4$ & $0.42\pm 0.41$ & $0.28$ & $0.74$ & $216$ &$1.19$x\\
\textsc{G-SAM}& $\mathbf{0.45 \pm 0.4}$ &$\mathbf{0.53 \pm 0.37}$&$\mathbf{0.29}$&$\mathbf{0.75}$&$\mathbf{17}$&$\mathbf{16.9}$x\\
\end{tabular}
    \caption{Segmentation models evaluation on Gemma3-4B with \textsc{P2} parameters; $t$ is the execution time in seconds per image, $v$ is the speed and \textsc{G-SAM} means \textsc{Grounded-SAM2}.}
    \label{tab:segmentation}
\end{table}

\paragraph{\textbf{\textsc{VizWiz-VQA-Grounding} Challenge}} The two best models selected in the previous steps are \textsc{Qwen2.5-VL-3B} and \textsc{Gemma3-4B-it}.  We present the results of these models on the entire \textsc{VizWiz-VQA-Grounding} validation set and on its challenge (the official test-set) in Table~\ref{tab:valtst-iou}. Note that although \textsc{Grounded SAM} outperforms all the other segmentation models, the IoU remains relatively low. 
\begin{table}[h!]
    \centering
    \begin{tabular}{l|c|c}\hline
       Models  & IoU & Dice \\\hline
       \multicolumn{3}{c}{\textsc{VizWiz-VQA-Grounding} validation set.}\\
        \hline
        Qwen2.5-VL-3B (Zero Shot BBox Grounding) & 0.33 & 0.43 \\
        Gemma3-4B-it (STEGO,P1) & 0.30 & 0.41 \\
        Gemma3-4B-it (SAM,P1)&0.38 & 0.44 \\
        Gemma3-4B-it (G-SAM, P2) & \textbf{0.45} & \textbf{0.52} \\
        Qwen2.5-VL-3B (G-SAM, P2) & 0.43 & 0.51 \\\hline
        \multicolumn{3}{c}{\textsc{VizWiz-VQA-Grounding} on-line challenge.}\\\hline
        Qwen2.5-VL-3B (Zero Shot BBox Grounding) & 0.33 & -\\   
        Gemma3-12B-it (SAM, P2)  & 0.40 & -\\
        Gemma3-4B-it (SAM, P2) &  0.41  & -\\
        Gemma3-4B-it (G-SAM, P2) & 0.43& -\\
        Qwen2.5-VL-3B (SAM, P2) & 0.43& -\\ 
        Qwen2.5-VL-3B (G-SAM, P2) & \textbf{0.45}& -\\
    \end{tabular}
    \caption{VizWiz Results, Dice is not reported by the challenge and BBox means Bounding Box. } 
    \label{tab:valtst-iou}
\end{table}

Figure~\ref{fig:sam-ex} shows the visual explanation process using \textsc{SAM} for the question: ``What is this?'' and the answer: ``It’s a bottle of `Trader Joe’s Milk Thistle Dietary Supplement' – 200 mg per capsule, 100 capsules''. The blue region on the image (Figure~\ref{fig:sam_explanation}) refers to the grounding mask. As the number of segments is too large ($89$), resulting in $2^{89}$ possible coalitions, we took the first 300 coalitions to generate the visual grounding explanation. Even with 300 samples, the generated explanation got the perfect cosine similarity (1.0) for the blue mask depicted in Figure~\ref{fig:sam_explanation}. The red regions are the features which got negative cosine similarity. The mask does not contain any noise as we observed with \textsc{STEGO}, and gives the best region of interest that contains all the details generated by the model. Note that the explosion of segments (i.e., a huge number of coalitions) was a clear limitation that we solved with \textsc{GroundedSAM2}.
\section{Conclusion and Future Work}
In this work, we successfully implemented VG-SHAP, a model-agnostic visual grounding explainability framework for VQA, and used it to assess the visual grounding capability of VLMs on images taken by VIP. It is worth noting that these images are particularly challenging due to poor quality (e.g., blurriness, out-of-focus, too bright, etc.). VG-SHAP combines external segmentation models with SHAP-based attribution methods, enabling us to assess visual grounding on VLMs without requiring access to internal attention maps or model architecture. Despite being effective at explaining black-box VLMs with a large number of parameters, this method has a notable limitation: the accuracy of the final grounding masks depends on the quality and granularity of the segments produced by an externally imposed segmentation model. 

SAM frequently produced an excessive number of fine-grained segments, which substantially increased computational overhead. Moreover,
some of these segments only partially covered the relevant visual region, leading to incomplete or misaligned visual grounding, even when the model produced a correct answer. This discrepancy resulted in lower IoU and Dice scores, highlighting a potential mismatch between segmentation granularity and semantic relevance. Although \textsc{Grounded-SAM} outperforms in terms of efficacy and visual grounding the other SAM variants, visual grounding scores remain low.

Our results suggest that VLMs tend to generate correct answers without grounding in the correct image segment. However, these results are constrained by the quality of the segmentation models. At this point, we cannot claim that these results are due to the VLM's poor grounding ability on VIP's images. In future work, we will need to further investigate visual grounding in VLMs for VIP, for instance, by exploring attention-based interpretability methods in open-source models and by conducting correlation analysis with respect to VG-SHAP.


\bibliographystyle{ACM-Reference-Format}
\bibliography{vlmeval}

\appendix
\section{Prompts}
\label{app:prompt}

In the following we present the prompt used in the experiments: Segmentation Models Evaluation and VizWiz-VQA-Grounding Challenge.

{\texttt{You are an assistant for visually impaired users. Provide a very specific response that accurately answers the question without any extra explanation based on the image. If the image is too blurry to answer reliably, respond with `unanswerable'.}} 

\section{Qualitative Analysis}
\begin{figure}[h!]
\centering
\includegraphics[width=0.48\textwidth]{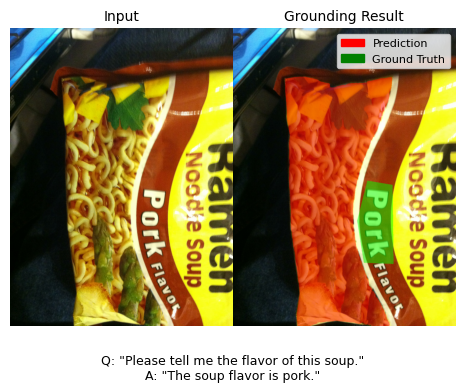}
\caption{
Representative visual grounding example from VizWiz using \textsc{Gemma3-4B-it} and \textsc{G-SAM}. The model correctly answers the question (“The soup flavor is pork”), but the predicted grounding mask (red) does not align with the ground-truth annotation (green).
}
\label{fig:qualitative_grounding}
\end{figure}

\textbf{Qualitative Visualizations and error analysis.}
We inspected several visual grounding outputs to identify common failure modes. We observed that VLMs may generate factually correct answers while grounding on incomplete or irrelevant regions, suggesting a mismatch between answer generation and visual evidence. In challenging VizWiz images (e.g., blurry, occluded or poorly illuminated scenes), segmentation errors often propagated to the final grounding mask, especially with fine-grained segmentations such as \textsc{SAM}. In some cases, relevant objects were fragmented into multiple regions, leading the explainer to select only a partial visual cue. 
Figure~\ref{fig:qualitative_grounding} illustrates a representative example from VizWiz using \textsc{Gemma3-4B-it} and \textsc{Grounded-SAM2}. The model produces a correct answer, while the predicted grounding mask does not align with the ground-truth annotation.

\end{document}